\documentclass[a4paper]{svproc}

\usepackage{makeidx}         
\usepackage{graphicx}        
\usepackage{multicol}        
\usepackage[bottom]{footmisc}

\usepackage{newtxtext}       %
\usepackage[varvw]{newtxmath}       
\usepackage{xspace}
\usepackage{multirow}
\usepackage{diagbox}
\usepackage{subfigure}
\usepackage{color}

\usepackage{array}
\newcolumntype{C}[1]{>{\centering\arraybackslash}p{#1}}

\makeindex             

\DeclareMathAlphabet{\mathcal}{OMS}{cmsy}{m}{n}

\def\I{\ensuremath{\mathbb{I}}}

\def\E{\ensuremath{\mathcal{E}}}

\newcommand{\Loss}{\mathcal{L}}

\newcommand{\demos}{\mathbf{X}}
\newcommand{\traj}{\mathbf{\Xi}}
\newcommand{\numberofdemos}{j}
\newcommand{\numberofpoints}{n}

\newcommand{\pose}{\mathbf{\xi}}

\newcommand{\gmmparam}{\theta}
\newcommand{\deltaparam}{\Delta\gmmparam}

\newcommand{\pparam}{\phi}

\newcommand{\policy}{\pi_{\pparam}}

\newcommand{\obs}{\mathbf{o}}

\renewcommand{\[}{\begin{equation}}
\renewcommand{\]}{\end{equation}}

\renewcommand{\eqref}[1]{Eq.~(\ref{#1})}

\def\ourmodel{KIS-GMM\xspace}

\begin{document}




\mainmatter
\title{Robot Skill Generalization via Keypoint Integrated\\Soft Actor-Critic Gaussian Mixture Models}
\titlerunning{Robot Skill Generalization via \ourmodel{}s}

\author{Iman Nematollahi$^{*1}$ \and Kirill Yankov$^{*1}$ \and Wolfram Burgard$^{2}$ \and Tim Welschehold$^{1}$}
\authorrunning{Iman Nematollahi$^{*}$ \and Kirill Yankov$^{*}$ \and Wolfram Burgard \and Tim Welschehold}

\institute{$^{1}$ Department of Computer Science, University of Freiburg, Germany\\
$^{2}$ Department of Engineering, University of Technology Nuremberg, Germany\\
$^{*}$ Equal contribution}

\toctitle{Lecture Notes in Computer Science}
\tocauthor{Authors' Instructions}
\maketitle

\sloppy
\begin{abstract}
\vspace{-0.2cm}
A long-standing challenge for a robotic manipulation system operating in real-world scenarios is adapting and generalizing its acquired motor skills to unseen environments. We tackle this challenge employing hybrid skill models that integrate imitation and reinforcement paradigms, to explore how the learning and adaptation of a skill, along with its core grounding in the scene through a learned keypoint, can facilitate such generalization. To that end, we develop Keypoint Integrated Soft Actor-Critic Gaussian Mixture Models (\ourmodel{}) approach that learns to predict the reference of a dynamical system within the scene as a 3D keypoint, leveraging visual observations obtained by the robot’s physical interactions during skill learning. Through conducting comprehensive evaluations in both simulated and real-world environments, we show that our method enables a robot to gain a significant zero-shot generalization to novel environments and to refine skills in the target environments faster than learning from scratch. Importantly, this is achieved without the need for new ground truth data. Moreover, our method effectively copes with scene displacements.\looseness=-1
\footnotetext{\url{http://kis-gmm.cs.uni-freiburg.de}}


\end{abstract}

\section{Introduction}
\label{sec:introduction}
\vspace{-0.2cm}
Robot skill generalization remains a long-standing challenge in the field of robotic manipulation, mainly due to the stochastic nature and inherent variability of unstructured real-world environments. For effective and autonomous operation, a robot must excel not only in acquiring new motor skills, but also in adapting and generalizing these skills to unseen and evolving contexts. However, this necessity often stands in contrast with the common practice of learning robot skills in imitation and reinforcement learning settings, which typically confine a robot to operate optimally within the rigid boundaries of its training environment. Such methods frequently stumble when faced with novel environments where the distribution of observations diverges from the one encountered during training. Moreover, the lack of adaptability intrinsic to most robot skill models imposes a significant constraint, hindering the robot's capacity to refine its learned skills within the target environment.\looseness=-1

\begin{figure*}[t]
    \centering
    \vspace{-0.3cm}
    \includegraphics[scale=0.655]{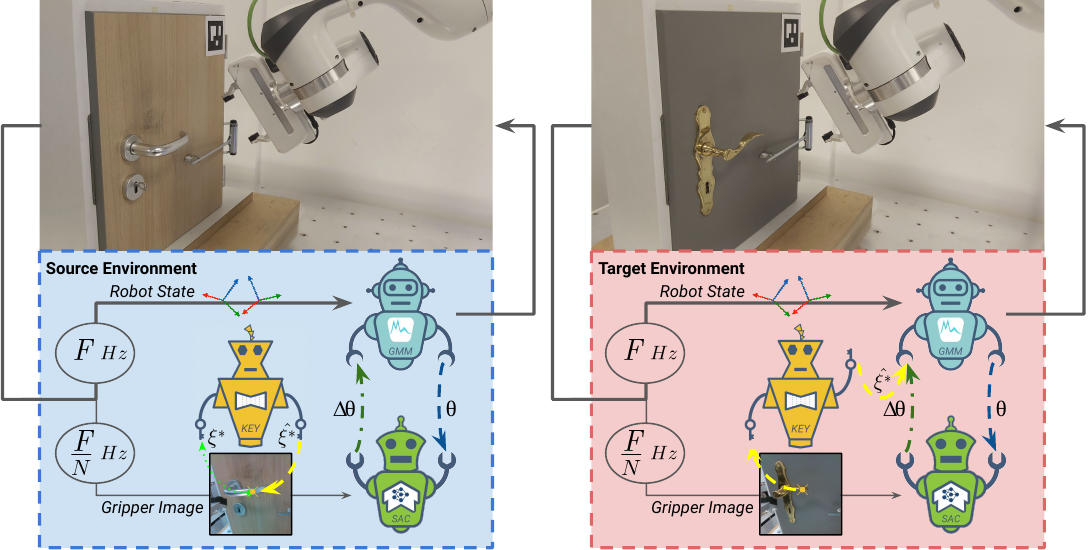}
    \vspace{-0.3cm}
    \caption{\ourmodel{} employs a hybrid model to acquire adaptive and generalizable robot skills. Initiating in a source environment (illustrated on the left), where the robot has access to few skill demonstrations and a pivotal keypoint ($\pose^{*}$) of the skill in the scene, the Gaussian Mixture Model (GMM) agent learns a dynamical system from these demonstrations and controls the robot at high-frequency \textit{F}. The Soft Actor-Critic (SAC) and Keypoint (KEY) agents, operating at lower frequencies, respectively learn to refine skills for adaptability and predict a 3D keypoint for generalization to novel environments. In transitioning to a target environment (illustrated on the right), where neither demonstrations nor ground-truth keypoints are available, KIS-GMM effectively predicts the essential keypoint ($\hat{\pose^{*}}$) for grounding the skill in the novel scene. This framework allows the seamless execution of skills, initially learned in a source environment, within unseen target settings.\looseness=-1}
    \label{fig:cover}
    \vspace{-0.1cm}
\end{figure*}
Numerous efforts have been made to address the need for generalizable and adaptive robot skills, yet predominantly addressed them as separate challenges. On the one hand, attempts to construct generalizable robot skills have primarily focused on enabling robots to learn object-centric and task-specific visual representations, fostering a higher readiness for their transferability to diverse yet structurally similar environments~\cite{jang2022bc,hartz23bask,gao2022k}. However, the lack of adaptability in these methods hinders performance refinement when initial generalizations fail. On the other hand, endeavors to create adaptive robot skills adopt a strategy that confines the refinement of a learned skill policy exclusively to its initial training environment~\cite{nematollahi2022robot,nair2022learning,ziesche2023wasserstein}. Nonetheless, while both pursuits are crucial, there has been limited exploration into the potential synergy of these two key aspects.\looseness=-1

\vspace{-0.2cm}
In this paper, we step towards closing this gap by proposing the Keypoint-Integrated Soft Actor-Critic Gaussian Mixture Models (\ourmodel{}, see Figure~\ref{fig:cover}). \ourmodel{}s employ a hybrid skill model that begins by learning a trajectory-based Gaussian mixture dynamical system from demonstrations. This learned skill is subsequently refined through physical interactions of a soft actor-critic reinforcement learning agent within its environment. During this refinement phase, \ourmodel{}s harness the robot's observations to predict a 3D keypoint corresponding to the dynamical system's reference point in the scene, empowering it to generalize to structurally similar but previously unseen environments. This is achieved by transforming the motion model to the reference frame given by the predicted keypoint in new settings. Through comprehensive evaluations, we demonstrate that our \ourmodel{} equips a robot with the ability for remarkable zero-shot generalization in unseen environments. It expedites skill refinement in the target environment, outpacing the learning-from-scratch approach, and does so without the need for the new environment’s ground truth data. Additionally, our model effectively handles scene displacements. We showcase \ourmodel{}'s capability to successfully open four visually distinct doors and drawers in the real world, having been initially trained on just one of each, respectively.\looseness=-1

\section{Related Work}
\label{sec:related_work}
A fundamental challenge for autonomous agents in unstructured real-world scenarios is to adapt and generalize their skill set to manipulate objects in dynamic, unfamiliar environments~\cite{cui2021toward}. For skill generalization, the focus is on \textit{where} to apply a previously learned skill in a new, unseen environment. In contrast, skill adaptation emphasizes \textit{how} to refine the robot's skill for optimal performance. Extensive research in robotics and computer vision has delved deeply into addressing either adaptability or generalization challenges.\looseness=-1

\textbf{Robot Skill Generalization:}
A core objective of machine learning is to distill abstract information from limited training data that can be generalized to unseen observations, often leveraging similarities in the training data to formulate an inductive bias~\cite{mitchell1980need}. In recent years, deep learning has facilitated the zero-shot and few-shot generalization of robotic skills within an imitation learning paradigm~\cite{jang2022bc,finn2017one,james2018task,yu2018one,dasari2021transformers,zhou2019watch,huang2019neural}. In robot skill learning, a common approach for incorporating inductive bias is through learning the visual representation of data. Various methods exist to derive such generalizable visual representations. This includes generative models like Variational Autoencoders (VAEs)~\cite{kingma2013auto} and Generative Adversarial Networks (GANs)~\cite{NIPS2014_5ca3e9b1} or object-centric models such as video prediction~\cite{finn2016unsupervised,nematollahi2022t3vip}, image or point cloud segmentation~\cite{oquab2023dinov2,kappeler2023few,xie2021unseen,byravan2017se3,nematollahi2020hindsight}, and keypoint detection~\cite{hartz23bask,gao2022k,manuelli2019kpam,chen2021unsupervised,vecerik2021s3k} networks. While robot skills using these generalization techniques can manipulate objects in novel environments, they often lack an adaptation mechanism for failed generalizations. In contrast, our skill model refines its keypoint prediction in response to sub-optimal generalizations in new scenes.

\textbf{Robot Skill Adaptation:}
While robot skill adaptation techniques have traditionally leaned on end-to-end deep reinforcement learning~\cite{ibarz2021train} and policy search~\cite{deisenroth2013survey}, there is a growing interest toward hybrid skill models aiming for better sample efficiency~\cite{nematollahi2022robot,nair2022learning}. These models begin by learning robot motion policies from supervised data and later refine them in the environment through reinforcement learning. While some approaches overlook the intrinsic policy structure during adaptation, either by optimizing policy parameters~\cite{lee2022offline,rajeswaran2018learning} or adding residual action at each time step~\cite{silver2018residual,johannink2019residual,hoppe2020sample}, others preserve the geometric structure of the initial policy~\cite{nematollahi2020hindsight,ziesche2023wasserstein}. In our previous study~\cite{nematollahi2022robot}, we introduced a framework to learn robot skills using Gaussian Mixture Models~\cite{khansari2011learning,pastor2009learning,figueroa2018physically} and subsequently adapt them in their trajectory distribution space using a Soft Actor-Critic agent~\cite{haarnoja2018soft} that interacts with the environment. While these approaches can adapt robot policies and enhance skill performance in the initial training environment, they struggle to generalize in new settings with different visual appearances, even if the skill goals remain similar.\looseness=-1

\vspace{-0.4cm}
\section{Problem Formulation}
\label{sec:problem}
\vspace{-0.3cm}
We consider the motion of a robot to be driven by a dynamical system regulated by a set of first-order differential equations, conditioning the robot's velocity on the current and target states~\cite{khansari2011learning}. Through this framework, we define a robot skill as a trajectory map, indicating the requisite motion to reach a target state from any starting state. Our goal is first to master a skill in one environment and subsequently generalize its application to unseen environments. This generalization is achieved by transferring the dynamical system's reference point, while offering the flexibility to adapt the skill in the new domain if needed.\looseness=-1

To this end, we regard robot skills as 3D trajectories. We represent a trajectory by $\traj = \langle \pose_1, \dots,\pose_n \rangle$, where $\pose_i \in \mathbb{R}^d$ is the robot's geometric pose. Importantly, each $\pose_n$ can exemplify a dynamical system's reference point, represented by $\pose^{*}$. Furthermore, each rollout of a robot skill prompts a collection of observations, denoted as $\mathcal{O} = \langle \obs_1, \dots,\obs_\numberofpoints \rangle$, captured from the environment through the robot's wrist camera. To learn a generalizable adaptive robot skill, we first need a series of demonstrations taken from one single environment, represented as $\demos = \langle \traj_1, \dots,\traj_\numberofdemos \rangle$, with each demonstration exemplifying the geometric course of the respective skill. Our initial goal is to learn a parametric skill model, denoted by $f_{\theta}(\pose - \pose^{*}) = \dot{\pose}$, which, conditioned on the dynamical system's reference point, derives the robot's next desired velocity given the current pose. Subsequently, we aim to master the learned skill in the source environment by adapting its trajectory parameters $\theta$ through a refinement policy $\policy(\pose, \obs) = \Delta\theta$, considering the robot's poses and observations in a sparse task completion reward setup.\looseness=-1

Seeking to generalize our skill model to structurally similar but unseen environments, our approach involves training a 3D keypoint detector, represented as $K_{\omega}(\obs) = \hat{\pose^{*}}$, during the skill refinement phase in the source environment. The keypoint detector leverages the wrist camera observations to predict the dynamical system's reference point. By transferring this representation to unseen environments, the robot can operate in novel settings. If the skill’s performance in the target environment proves unsatisfactory, we retain the capability to refine both the skill and the keypoint detector specifically for that environment, utilizing sparse task completion rewards garnered within this novel context.\looseness=-1

\section{\ourmodel{}}
\label{sec:approach}
We introduce the Keypoint Integrated Soft Actor-Critic Gaussian Mixture Models (\ourmodel{}), a robot learning framework specifically crafted for fostering adaptive and generalizable robot skills. This framework is segmented into two distinct phases. In the first phase, a robot skill is acquired by learning a parameterized dynamical system from a few demonstrations within a single source environment. This stage masters the skill not only by adapting dynamical system parameters through physical interactions, high-dimensional sensory data, and sparse rewards but also by deriving a pivotal keypoint crucial for grounding the skill in the scene. In the second phase, the dynamical system is generalized to unseen environments that share structural similarities to the source environment by transferring the learned keypoint, conducting any necessary refinements to either the skill policy or keypoint detection if required.\looseness=-1

\vspace{-0.3cm}

\subsection{Mastery of a Robot Skill in a Source Environment}
We employ the SAC-GMM hybrid algorithm~\cite{nematollahi2022robot}, a fusion of dynamical systems and deep reinforcement learning, for an adaptive formulation of robot skills. Initial learning of a parametrized dynamical system stems from a mixture of Gaussians, derived from a handful of demonstrations (GMM agent). The Soft Actor-Critic framework (SAC agent) then exploits robot interactions with the source environment to refine this dynamical system further. Provided with the current state - including the latent observation and the robot state - and a reward for the previous action, the SAC agent adapts the original GMM for the following interactions. For detailed insights into SAC-GMM, refer to our prior work~\cite{nematollahi2022robot}.\looseness=-1

\begin{figure*}[t]
    \centering
    \includegraphics[scale=0.55]{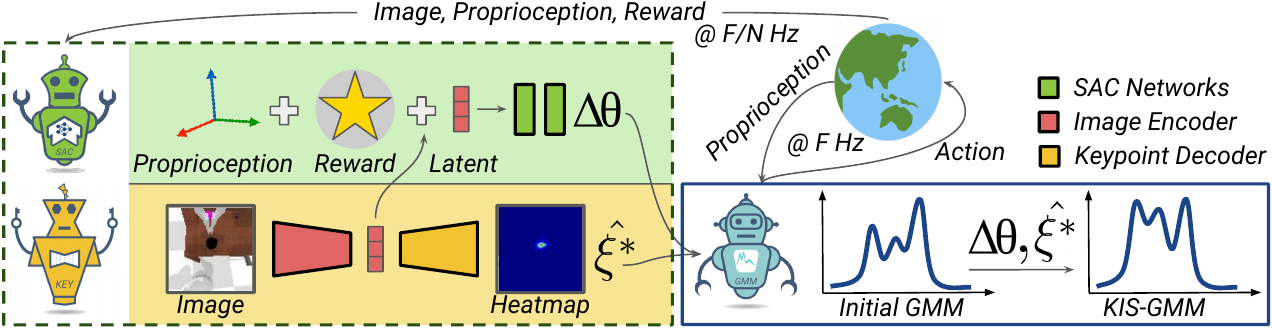}
    \vspace{-0.3cm}
    \caption{\ourmodel{} learns to adapt a robot skill and generalize it to unseen environments. Adaptation is achieved through parameter refinements ($\deltaparam$) utilizing physical interactions and sparse rewards. Generalization is facilitated by transferring a learned keypoint ($\hat{\pose^{*}}$) to novel environments.\looseness=-1}
    \label{fig:architecture}
\end{figure*}

We develop a 3D keypoint detector (KEY agent), that builds upon the SAC-GMM algorithm. Our keypoint detection approach draws inspiration from and resembles the kPAM framework~\cite{manuelli2019kpam}. During the skill refinement phase within the source environment, the Key agent utilizes the wrist camera’s observations to predict the 3D position of the dynamical system’s reference point in the frame of the wrist camera, represented as $\pose^{*} = (x_c, y_c, z_c)$. For clarity, within the source environment, the ground truth keypoint is designated as a specific point located on the surface of the object that is integral to the relevant skill. For example, in the case of door and drawer opening skills, this keypoint is strategically positioned on the handles. Our KEY agent employs image encoding to generate a 2D heatmap, facilitating the determination of the keypoint’s coordinates within the camera’s frame. Concurrently, a depth map is produced to capture the depth value of the corresponding keypoint. Together, these processes result in the determination of $\hat{\pose{*}} = (\hat{x_c}, \hat{y_c}, \hat{z_c})$. Moreover, a keypoint presence probability evaluates the uncertainty regarding the keypoint’s presence within the camera’s frame. Using the ground truth keypoint of the dynamical system, we train the Keypoint detector with average Euclidean distance ($\Loss_{ED}$) and the keypoint presence measure via binary cross entropy loss ($\Loss_{KP}$):\looseness=-1
\begin{equation*}
    \label{eq:kp_loss}
    \Loss_{ED}(\omega) = \frac{1}{N} \sum_{i=0}^{N} \sqrt{(\hat{\pose^{*}_{i}}-{\pose}^{*})^{2}}, \quad
    \Loss_{KP}(\omega) = -\frac{1}{N}\sum_{i=0}^{N}[y_i \log(\hat{y_i}) + (1-y_i) \log(1-\hat{y_i})].
\end{equation*}
To ensure that the keypoint detector effectively learns the crucial features associated with the ground truth keypoint, we employ image augmentation techniques during the training process. Specifically, we randomly obscure portions of the image pixels where the keypoint pixel is absent, enhancing the model's recognition of essential features contributing to the keypoint. During inference, we maintain a running average of the predicted keypoints for which the keypoint presence measure exceeds a predetermined threshold.\looseness=-1
\subsection{Generalization and Adaptation of a Robot Skill to Unseen Environments}
The ability of our skill model to learn the dynamical system's reference keypoint enables the transfer of this task-specific representation to structurally similar, yet previously unseen environments. Consequently, when faced with a new environment, our model is capable of grounding the skill in the new scene using the predicted keypoint. If performance in a target environment falls short of expectations, our model leverages sparse task completion rewards from that environment to fine-tune both the SAC and KEY agents. Since access to the ground truth reference point is unavailable in new environments, we adopt the running average of the predicted keypoints from the last three successful trajectories as a pseudo-label, enabling further refinement of the KEY agent.\looseness=-1

\section{Experimental Evaluation}
\label{sec:results}
\vspace{-0.3cm}
We evaluate the effectiveness of \ourmodel{} in acquiring adaptable and generalizable robot skills within simulated and real-world environments. Our experiments aim to investigate:
(i) the generalization ability of our robot skill model to structurally similar yet unseen environments, (ii) the time efficiency of refining a skill in a target environment versus learning it from scratch, and (iii) our model's capacity to handle scene displacements.\looseness=-1

\vspace{-0.4cm}
\subsection{Experimental Setup}
\begin{figure*}[t]
    \vspace{-0.4cm}
    \centering
    \includegraphics[scale=0.65]{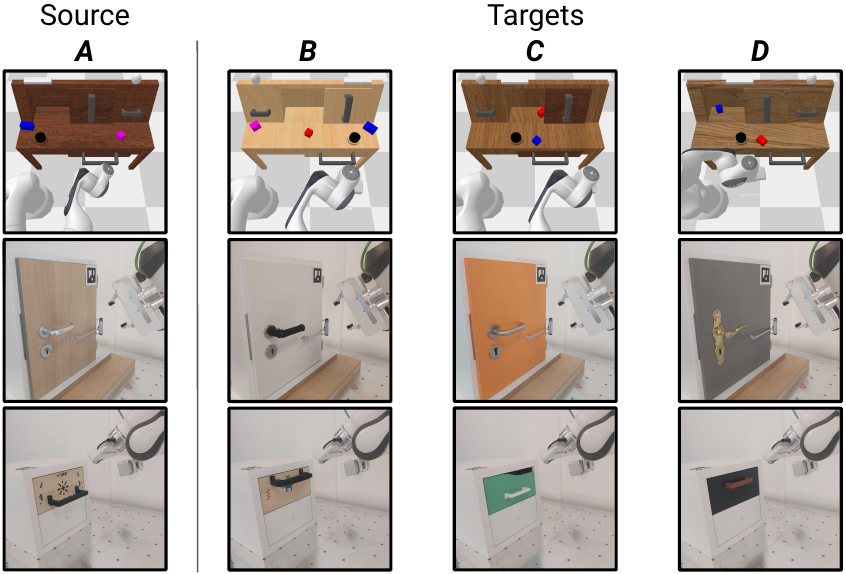}
    \vspace{-0.4cm}
    \caption{The top row displays four simulated CALVIN environments, while the second and third rows present four real-world door and drawer setups respectively. In all cases, robot skills are trained in source environment \textit{A} and then generalized to environments \textit{B}, \textit{C}, and \textit{D}.\looseness=-1}
    \label{fig:envs}
    \vspace{-0.2cm}
\end{figure*}
\vspace{-0.2cm}
We assess our approach in both simulated and real-world settings. In the simulation, we conduct our experiments within the CALVIN manipulation environments~\cite{mees2022calvin}. CALVIN provides four distinct environments, engineered to test the generalization of robot skills in structurally similar contexts. Each environment varies in texture and the positioning of static elements like drawers and buttons, see the first row of Figure~\ref{fig:envs}. In CALVIN, We pursue the mastery of two distinct robot skills - \textit{Drawer Opening} and \textit{Button Pushing}.\looseness=-1 

In the real-world experiment, we examine \textit{Door Opening} and \textit{Drawer Opening} skills. We utilize our 7-DOF Franka Emika Panda robot arm, equipped with a parallel gripper, to interact with four distinct doors and drawers. These vary in attributes like color, texture, handle design, and placement, as depicted in the second and third rows of Figure~\ref{fig:envs}. We obtain the robot's wrist camera observations using a FRAMOS Industrial Depth Camera D435e mounted on the gripper. We affix ArUco markers to doors and drawers, allowing us to identify when they are opened successfully and provide a corresponding reward. Additionally, to facilitate autonomous robot operation without human intervention, our doors and drawers are fitted with closing mechanisms. As a result, the doors and drawers close automatically when the robot lets go of the handle, initiating a new interaction episode.\looseness=-1

During every training and test episode, the robot's end effector's starting position is randomly sampled from a Gaussian distribution. Additionally, we add a uniform noise distribution to the skill's reference point in the scene. For all simulated and real-world setups, robot skills are trained within the source environment \textit{A}. Subsequently, we assess the generalization and adaptability of these learned skills within environments \textit{B}, \textit{C}, and \textit{D}.\looseness=-1

\subsection{Evaluation Protocol}
\vspace{-0.2cm}
We compare our \ourmodel{} against the following baselines:\looseness=-1
\begin{itemize}
  \item \textbf{GMM}: This baseline utilizes the same dynamical system learned from demonstrations in our approach. However, it lacks adaptability and awareness of the dynamical system's reference point in the target environments.
  \item \textbf{SAC-GMM (w/ GT)}~\cite{nematollahi2022robot}: This skill model utilizes robot interactions and observations for adaptability and refinement of the GMM baseline. Additionally, it benefits from access to ground-truth reference points in the target environments, which aids its performance in these contexts.
  \item \textbf{SAC-GMM}: This baseline is similar to the previous one but lacks access to ground truth in the target environments, making it a closer representation of real-world scenarios.
  \item \textbf{KEY-GMM}: Building on the GMM baseline, this model integrates a keypoint detector to predict the dynamical system's reference point in target environments. However, it still lacks adaptability.\looseness=-1
\end{itemize}
All quantitative results represent the mean success rate of each skill model over 100 trials, computed across three different random seeds. 

\vspace{-0.4cm}
\subsection{Experiments in Simulation}
\begin{table}[b]
\centering
\setlength\tabcolsep{1.8pt}
\begin{tabular}{ |l|C{1cm}|C{1cm}|C{1cm}|C{1cm}|C{1cm}|C{1cm}|C{1cm}|C{1cm}| } 
\hline
\multirow{3}{*} {\diagbox[width=3.1cm, height=1.1cm, innerleftsep=0.3em, innerrightsep=0.3em]{Model}{Robot\\Skill }} & \multicolumn{4}{c|}{Drawer Opening} & \multicolumn{4}{c|}{Button Pushing} \\
\cline{2-9}
& Source & \multicolumn{3}{c|}{Target} & Source & \multicolumn{3}{c|}{Target}  \\ 
\cline{2-9}
& \textit{A} & \textit{B} & \textit{C} & \textit{D} & \textit{A}  & \textit{B} & \textit{C} & \textit{D} \\ 
\hline
GMM & 73\% & 5\% & 4\% & 5\% & 88\% & 5\% & 6\% & 6\% \\ 
SAC-GMM (w/ GT)& \textbf{96}\% & 60\% & \textbf{96}\% & 93\% & 95\% & 39\% & 66\% & 84\% \\ 
SAC-GMM & 96\% & 6\% & 8\% & 9\% & 95\% & 9\% & 14\% & 16\% \\ 
KEY-GMM & 67\% & 42\% & 60\% & 61\% & 94\% & 40\% & 58\% & 79\% \\ 
\ourmodel{} (Zero-Shot) & 95\% & 52\% & 94\% & 81\% & \textbf{96}\% & 33\% & 57\% & 80\% \\
\ourmodel{} (Refined) & x & \textbf{93}\% & 94\% & \textbf{93}\% & x & \textbf{97}\% & \textbf{94}\% & \textbf{95}\%  \\ 
\hline
\end{tabular}
\vspace{0.1cm}
\caption{ Superior performance of \ourmodel{} over baseline models in generalizing and adapting \textit{Drawer Opening} and \textit{Button Pushing} skills, initially mastered in environment \textit{A}, to target environements \textit{B}, \textit{C} and \textit{D}.\looseness=-1}
\label{table:generalization}
\end{table}
\vspace{-0.1cm}
We begin by assessing our approach with the \textit{Drawer Opening} and \textit{Button Pushing} skills in CALVIN environments. Table~\ref{table:generalization} reports the quantitative results representing the accuracy with which each skill model generalizes to unseen target environments. Our \ourmodel{} is not only effective in achieving notable zero-shot generalization to unseen environments, but it is also capable of refining its performance in target environments. Remarkably, it accomplishes this refinement without requiring access to target environments' ground truth data. As a result, \ourmodel{} achieves an accuracy exceeding 90\% across all environments for both assessed skills. Upon investigating the performance of the SAC-GMM baselines, we find that while they efficiently refine the initial GMM model in source environments and can generalize to unseen environments given access to ground truth data, they fail in the more realistic scenario where access to such data is unavailable. Furthermore, while the KEY-GMM models exhibit results that come close to matching the zero-shot variations of \ourmodel{}, they lack the capability to refine their performance. This often culminates in them achieving sub-optimal accuracy rates.

As Table~\ref{table:generalization} illustrates, the zero-shot generalization of skills to environment \textit{B} can lead to subpar results, largely due to significant table texture changes. In contrast to KEY-GMM, our adaptive skill model demonstrates the capacity to refine its performance in the target environment when zero-shot generalization doesn't meet expectations. As depicted in Figure 2, the comparative timelines for skill refinement in a new domain versus mastering the skill from the ground up reveal a distinct advantage for our model. For both robot skills assessed, \ourmodel{} refines the skill significantly faster than the time it would take to learn it from scratch (approximately four times quicker), importantly, without requiring the ground truth reference point of the dynamical system in the target environment.\looseness=-1
\begin{figure*}[t]
    \vspace{-0.4cm}
    \centering
    \subfigure[Drawer Opening in Environment \textit{B}]{
        \includegraphics[width=0.45\textwidth]{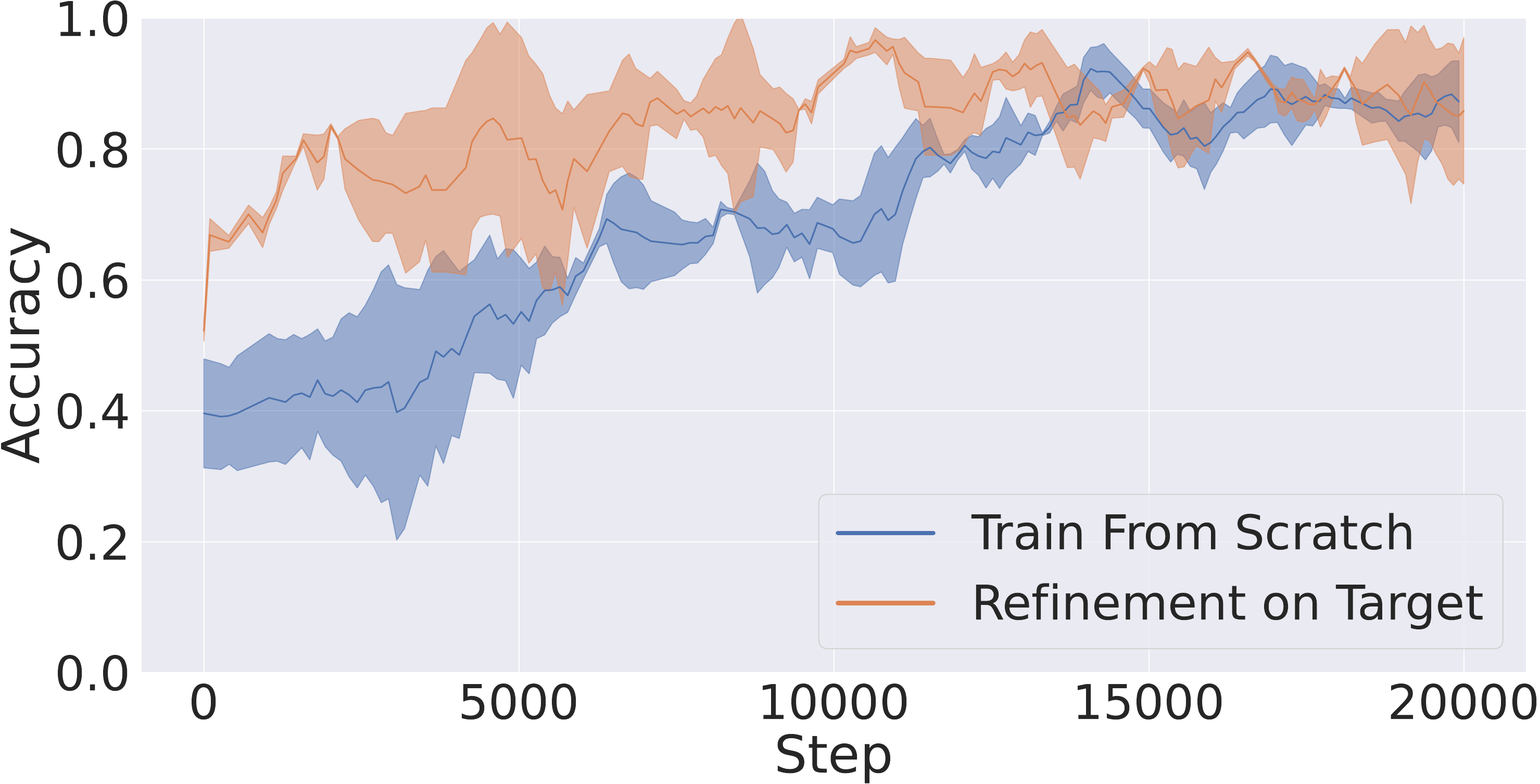}\label{fig:drawer_time}}
    \subfigure[Button Pushing in Environment \textit{B}]{
        \includegraphics[width=0.45\textwidth]{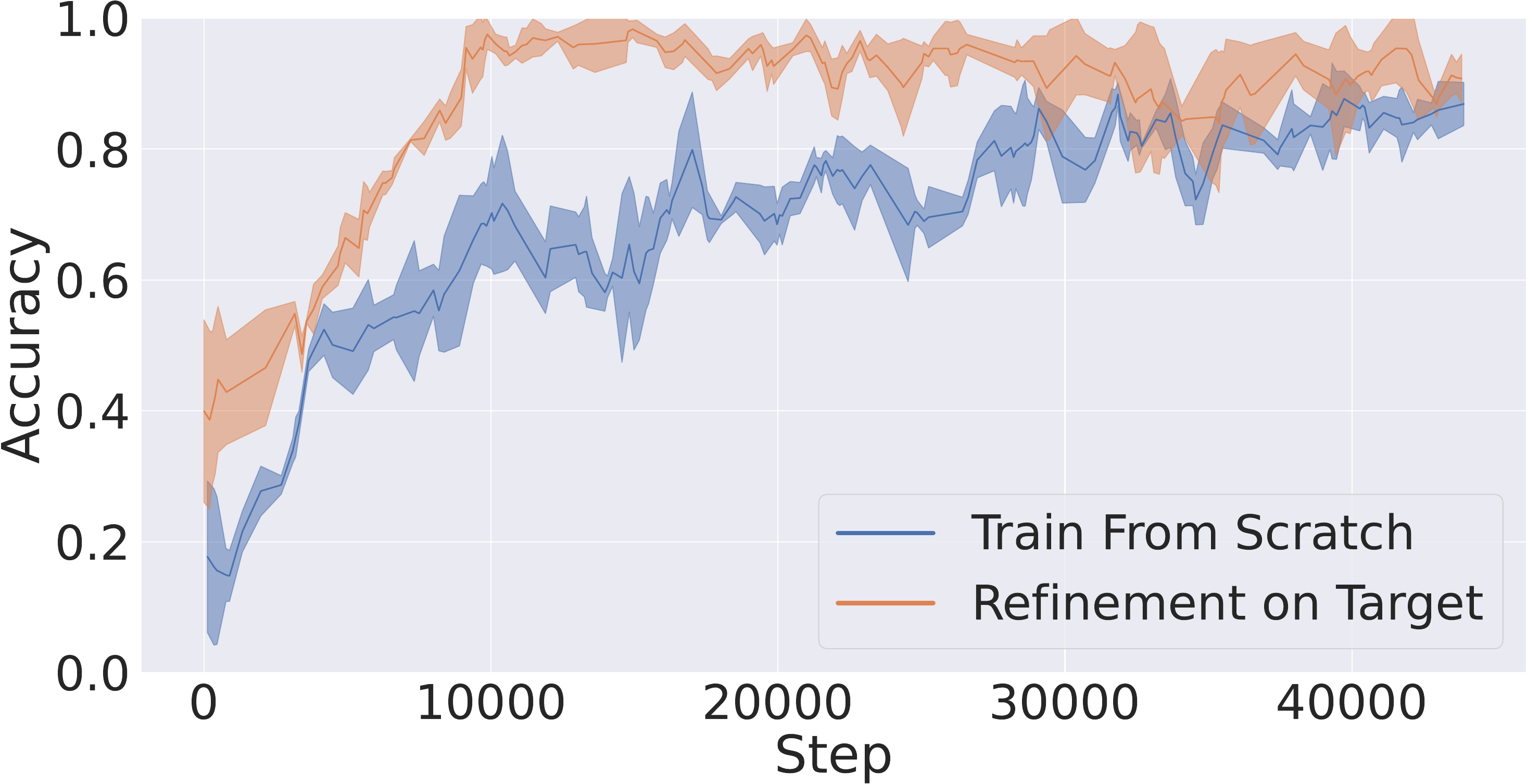}\label{fig:led_time}}
    \vspace{-0.4cm}
    \caption{\ourmodel{} swiftly refines both robot skills in the target environment \textit{B}, outpacing learning the skills from scratch, without even requiring ground truth reference points.}
    \label{fig:plots}
    \vspace{-0.2cm}
\end{figure*}

\begin{table}[b]
\centering
\setlength\tabcolsep{1.8pt}
\begin{tabular}{ |l|C{1.05cm}|C{1.05cm}|C{1.05cm}|C{1.05cm}|C{1.05cm}|C{1.05cm}|C{1.05cm}|C{1.05cm}|} 
\hline
\multirow{3}{*} {\diagbox[width=2cm, height=1.15cm, innerleftsep=0.3em, innerrightsep=0.3em]{Model}{Robot\\Skill }} & \multicolumn{4}{c|}{Drawer Opening / Environment \textit{A}} & \multicolumn{4}{c|}{Button Pushing / Environment \textit{A}} \\
\cline{2-9}
& Noise & \multicolumn{3}{c|}{Noisy} & Noise & \multicolumn{3}{c|}{Noisy}  \\ 
\cline{3-5}
\cline{7-9}
& Free & Easy & Medium & Hard & Free & Easy & Medium & Hard \\ 
\hline
GMM & 73\% & 23\% & 5\% & 5\% & 88\% & 35\% & 5\% & 1\% \\ 
SAC-GMM & \textbf{96}\% & 64\% & 3\% & 2\% & 95\% & 92\% & 25\% & 3\% \\ 
KEY-GMM & 67\% & 72\% & 65\% & 65\% & 89\% & 93\% & 90\% & 92\% \\ 
\ourmodel{} & 95\% & \textbf{97}\% & \textbf{94}\% & \textbf{97}\% & \textbf{96}\% & \textbf{96}\% & \textbf{92}\% & \textbf{95}\% \\
\hline
\end{tabular}
\vspace{0.1cm}
\caption{\ourmodel{} is effective in coping with scene displacements between training and evaluation stages, by employing a 3D keypoint detector for grounding its dynamical system.\looseness=-1}
\label{table:shift_result}
\end{table}
Our subsequent experiment investigates the adaptability of various skill models to scene displacements occurring between the training and evaluation phases. Specifically, we manipulate the relative positioning of the robot and tables and delineate three tiers of displacement difficulty for adaptation: ``\textit{Easy}'' signifies displacements from 0 to 10 centimeters, ``\textit{Medium}'' represents displacements from 10 to 20 centimeters, and ``\textit{Hard}'' encompasses displacements from 20 to 30 centimeters. All displacements are randomly sampled from a uniform distribution. As indicated in Table~\ref{table:shift_result}, the adaptive SAC-GMM model, while outperforming the GMM baseline in ``\textit{Easy}'' displacement scenarios, falls short in managing the more challenging ``\textit{Medium}'' and ``\textit{Hard}'' scenarios. This limitation primarily stems from these models' inability to ground their motion models to the displaced scene by predicting reference keypoints. Both KEY-GMM and \ourmodel{} skill models demonstrate a capability to handle displacement noise effectively due to their training in detecting the 3D keypoint corresponding to the dynamical system's reference point within the environment. Consequently, their performance remains robust despite considerable displacements between the training and evaluation stages.\looseness=-1

\subsection{Experiments in Real-World}
Table~\ref{table:real_experiments} presents the accuracy results of the \textit{Door Opening} and \textit{Drawer Opening} skill models as they generalize to previously unseen target settings. For both of these skills, the refinement phase spanned roughly five hours of physical interactions, equating to approximately 1,000 episodes, during which the SAC and KEY agents were actively trained. These results echo the outcomes observed in the simulated environments, underscoring the robustness and general applicability of our findings. \ourmodel{} demonstrates remarkable zero-shot generalization and adaptability in the real-world setting, outperforming all baseline models. GMM and SAC-GMM models, lacking awareness of the dynamical system's reference point in target environments, struggle to anchor the skill to these new contexts, leading to an inability to generalize the learned skill effectively. Additionally, though the KEY-GMM models can generalize robot skills to unfamiliar settings, their inability to adapt these skills often leads to subpar performance.
Our \ourmodel{} model consistently achieves an accuracy rate of over 80\% across all environments for both evaluated skills, with the sole exception of the \textit{Door Opening} skill in environment \textit{D}. 
Notably, while our KEY agent adeptly identifies the door handle and anchors the skill within the scene, the door handle in environment \textit{D} is designed with a slippery surface, inherently altering its point of maximum leverage compared to other environments. Recognizing this nuance, we undertook further refinement of the door opening skill model specifically for environment \textit{D}. After approximately an hour of refinement ($\sim$200 episodes), the accuracy of our model increased to 83\%.

\begin{table*}[t]
\vspace{-0.4cm}
\centering
\setlength\tabcolsep{1.8pt}
\begin{tabular}{ |l|C{1cm}|C{1cm}|C{1cm}|C{1cm}|C{1cm}|C{1cm}|C{1cm}|C{1cm}| } 
\hline
\multirow{3}{*} {\diagbox[width=3.1cm, height=1.1cm, innerleftsep=0.3em, innerrightsep=0.3em]{Model}{Robot\\Skill }} & \multicolumn{4}{c|}{Real World Door Opening} & \multicolumn{4}{c|}{Real World Drawer Opening} \\
\cline{2-9}
& Source & \multicolumn{3}{c|}{Target} & Source & \multicolumn{3}{c|}{Target}  \\ 
\cline{2-9}
& \textit{A} & \textit{B} & \textit{C} & \textit{D} & \textit{A}  & \textit{B} & \textit{C} & \textit{D} \\ 
\hline
GMM & 56\% & 11\% & 9\% & 5\% & 67\% & 6\% & 17\% & 8\% \\ 
SAC-GMM & 93\% & 5\% & 13\% & 7\% & \textbf{96}\% & 6\% & 19\% & 15\% \\ 
KEY-GMM & 49\% & 44\% & 55\% & 23\% & 65\% & 59\% & 52\% & 34\% \\ 
\ourmodel{} (Zero-Shot) & \textbf{93}\% & \textbf{84}\% & \textbf{88}\% & 41\% & 94\% & \textbf{82}\% & \textbf{89}\% & \textbf{86}\% \\
\ourmodel{} (Refined) & x & x & x & \textbf{83}\% & x & x & x & x  \\ 
\hline
\end{tabular}
\vspace{0.1cm}
\caption{KIS-GMM effectively masters the task of opening four distinct doors and drawers in real-world settings, after being trained initially on just a single door and drawer (source environment \textit{A}).\looseness=-1}
\vspace{-0.4cm}
\label{table:real_experiments}
\end{table*}

\vspace{-0.4cm}

\section{Conclusions}
\label{sec:conclusion}
\vspace{-0.2cm}
In conclusion, our novel KIS-GMM framework promotes robot skills that are both adaptable and generalizable, emphasizing their mutual importance for superior performance across diverse scenarios. Using the robot's observations from the skill refinement phase, our model learns to identify a 3D keypoint representation that grounds the skill's dynamical system in a novel scene. Comprehensive experiments in simulation and real-world highlight the efficacy of our proposed skill model in several key areas: 1) generalizing robot skills to unseen environments and subsequently adapting them within these new domains, 2) reducing skill refinement time in a target domain over learning from scratch, even without requiring ground truth data, and 3) handling scene displacement between training and evaluation stages. A compelling extension of our research would be to explore enhancing the sample efficiency of our skill model even further. This could be achieved by incorporating multiple keypoints that compactly define the robot's task, and by granting the refinement agent access to such keypoints.
\looseness=-1

\bibliographystyle{ieeetr}
\bibliography{sources}
\end{document}